\documentclass[twoside,leqno,twocolumn]{article}

\usepackage[letterpaper]{geometry}

\usepackage{ltexpprt}
\usepackage{hyperref}
\usepackage{subfigure}
\usepackage{graphicx}
\usepackage{makecell}
\usepackage{wrapfig}
\usepackage{booktabs}
\usepackage{amsmath}
\usepackage{amsfonts}

\begin{document}

\newcommand\relatedversion{}
\renewcommand\relatedversion{\thanks{The full version of the paper can be accessed at \protect\url{https://arxiv.org/abs/1902.09310}}} 

\title{Dual-stage Flows-based Generative Modeling for Traceable Urban Planning}
\author{Xuanming Hu\footnotemark[1], Wei Fan\footnotemark[2], Dongjie Wang\footnotemark[3], Pengyang Wang\footnotemark[2] \footnotemark[5], Yong Li\footnotemark[4], Yanjie Fu\footnotemark[1] \footnotemark[5]}


\maketitle

\footnotetext[1]{School of Computing and Augmented Intelligence, Arizona State University. Email: \{solomonhxm, yanjie.fu\}@asu.edu}
\footnotetext[2]{Department of CIS, SKL-IOTSC, University of Macau. Email: \{pywang, weifan\}@um.edu.mo}
\footnotetext[3]{Department of Computer Science, University of Central Florida. Email: dongjie.wang@ucf.edu}
\footnotetext[4]{Department of Electronic Engineering, Tsinghua University. Email: liyong07@tsinghua.edu.cn}
\footnotetext[5]{Corresponding authors.}

\begin{abstract}
  Urban planning, which aims to design feasible land-use configurations for target areas, has become increasingly essential due to the high-speed urbanization process in the modern era. However, the traditional urban planning conducted by human designers can be a complex and onerous task. Thanks to the advancement of deep learning algorithms, researchers have started to develop automated planning techniques. While these models have exhibited promising results, they still grapple with a couple of unresolved limitations: 1) Ignoring the relationship between urban functional zones and configurations and failing to capture the relationship among different functional zones. 2) Less interpretable and stable generation process. To overcome these limitations, we propose a novel generative framework based on normalizing flows, namely Dual-stage Urban Flows (DSUF) framework. Specifically, the first stage is to utilize zone-level urban planning flows to generate urban functional zones based on given surrounding contexts and human guidance. Then we employ an Information Fusion Module to capture the relationship among functional zones and fuse the information of different aspects. The second stage is to use configuration-level urban planning flows to obtain land-use configurations derived from fused information. We design several experiments to indicate that our framework can outperform compared to other generative models for the urban planning task.
\end{abstract}
\vspace{-0.3cm}
\section{Introduction}
\vspace{-0.1cm}
The fast urbanization process in the modern era poses substantial challenges to public administration due to the escalating need for urban construction~\cite{jarah2019urbanization}. Consequently, the importance of urban planning has grown significantly in the effective management of modern cities. Feasible planning can potentially enhance the convenience of citizens' lives, drive sustainable economic expansion, and safeguard the environment. Given these circumstances, urban planning has emerged as a prominent subject of research in recent times~\cite{levy2016contemporary, saaty2017rethinking}. 

\begin{figure}[t]
\centering
\includegraphics[width=3in]{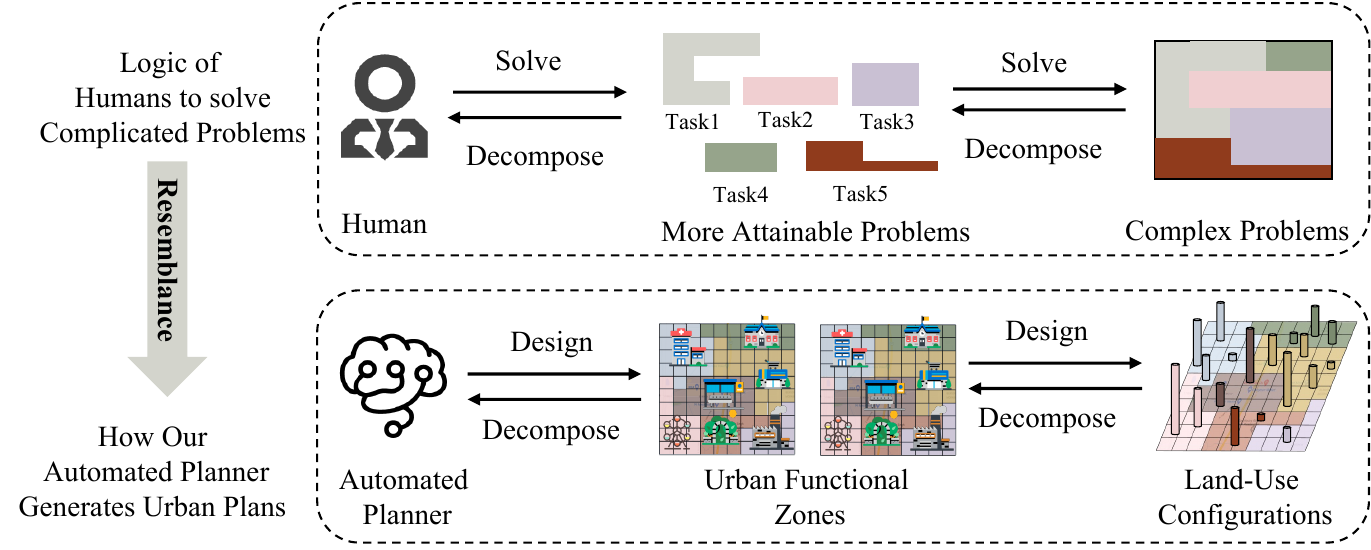}
\vspace{-2mm}
\caption{The proposed automated planner can resemble the logic of human and generate urban functional zones and land-use configurations step by step.}
\vspace{-6mm}
\label{intro}
\end{figure}

The traditional urban planning undertaken by human planners can be intricate and demanding due to numerous socio-economic factors to be considered such as surrounding traffic conditions, environmental issues, and business synergy~\cite{niemela1999ecology,levy2016contemporary}.
Consequently, it is imperative to develop a time-saving method to re-investigate the urban planning process. Recently, with the development of deep learning generation methods, researchers have started to construct deep learning models tailored for urban planning tasks~\cite{wang2021deep,shen2020machine, wang2023automated, wang2023towards}.~\cite{wang2021deep} utilized CLUVAE with variational Gaussian embedding mechanism to integrate human guidance as well as surrounding context information for the automated urban planning task.~\cite{shen2020machine} employed a GAN-based generative model to create urban design plans with certain conditions of the urban area. Although existing works can generate auspicious results, they have the following limitations: 1) Neglecting the connection between the urban functional zone and the configuration, and failing to capture the relationship between diverse functional zones. 2) Less interpretable, trustworthy, and stable generation process.

To tackle these problems, we propose \textit{\textbf{Dual-stage Urban Planning Flows}} (DSUP Flows) framework, which can consider the relationship of urban functional zone and configuration as well as capture the planning dependencies, and also outperform with a more interpretable and stable generation process. For complex issues, humans try to break down the final goal into several smaller objectives~\cite{turner1999handbook}.
Similarly, we can decompose the urban planning process into two more attainable stages. Initially, we draw a rough sketch for the target area. Based on it, we generate detailed urban solutions. So our proposed framework can capture the hierarchical dependencies between urban functional zones and land-use configurations. 

Planning dependencies among functional zones, i.e., semantic correlation and geographical relationship, can also be crucial for urban planning.
To catch the dependencies, we employ the Information Fusion Module consisting of the convolution extraction block and the semantic projection block.
Geographical correlation reflects the location relationship of urban solutions. For instance, 
high-polluted buildings such as industrial facilities should be distant from the tourist attractions.
To capture the geographical relationship, we utilize a convolution extraction block to extract the representation of urban functional zones.
For the semantic correlation, urban functionalities such as low crime rates, and higher economic effects should be considered during the process of urban planning.
For example, 
if the economic benefit for the target area is required, adequate transportation facilities which are needed to attract consumers should be generated for the commercial zone.
We use a semantic projection block to capture the semantic correlation and fuse it with geographical information.

Compared to the "Black Box generation process" of GAN or VAE-based models, the flows network can reveal the reconstruction process step by step which can make it more trustworthy. Also, flows-based model only needs to include the negative log-likelihood function to train which increases the stability. It also employs a fully invertible structure that can map any urban solutions to the latent flow space 
and ensure accurate 
reconstruction~\cite{lugmayr2020srflow}. This allows us to develop a powerful manipulation technique for controlling the generation process. We design two novel Normalizing Flows-based models for the consistency of the framework, namely zone-level urban planning flows and configuration-level urban planning flows.

The main contributions of this paper can be summarized as follows:
(1) A novel DSUF framework is constructed for dual-stage urban planning. We generate urban functional zones in the first stage and obtain the land-use configurations in the second stage. A novel Information Fusion Module
is utilized to catch planning dependencies and connect these two stages.
(2) We design two novel normalizing flows to generate urban functional zones and land-use configurations, namely zone-level urban planning flows and configuration-level urban planning flows.
(3) We conduct several experiments on data from Beijing, and the
empirical results demonstrate the effectiveness and superiority of our DSUP Flows.
\vspace{-0.3cm}
\section{Related Work}
\vspace{-0.1cm}
\noindent\textbf{Urban Planning.}
Urban planning is becoming increasingly crucial in the contemporary day due to the growing demand for building sustainable smart cities ~\cite{papa2016smart}. For the purpose of developing feasible urban planning solutions, human experts usually make great effort to contemplate several constraints, such as socioeconomic factors and environmental protocol ~\cite{bathrellos2012potential, ratcliffe2004urban}. For instance, ~\cite{niemela1999ecology} clarifies the connection between urban planning and urban ecosystem. However, the structure of modern cities can be really complex, as a result, it is time-consuming for humans to take all constraints into account and propose a reasonable urban planning solution. Due to the incredible success of deep learning models in other domains, researchers have recently resorted to building deep generative learning models for the challenge of urban planning ~\cite{wang2021deep,wang2020reimagining, wang2023human}. ~\cite{wang2020reimagining} developed LUCGAN to generate land-use configurations for empty areas based on surrounding contexts. 
Compared to these previous works, our proposed DSUF framework involves a dual-stage generation process and an informative data fusion process. It has the ability to generate more decent configurations based on surrounding contexts and human instruction.

\noindent\textbf{Normalizing Flows.} 
Normalizing flows can be regarded as a set of invertible functions which aims to model an unrevealed distribution~\cite{dinh2014nice,dinh2016density}. Previous works have demonstrated theoretically and empirically that normalizing flows can perform well in several generation tasks such as image generation ~\cite{kingma2018glow,papamakarios2017masked} and text generation ~\cite{ma2019flowseq}. However, it is unrealistic to control the generating results of normalizing flows. Therefore, conditional normalizing flows is proposed to solve this problem ~\cite{papamakarios2017masked}. For instance, ~\cite{lugmayr2020srflow} used SRFlow to model the complex conditional distribution of high-resolution images given the low-resolution images. In contrast with preceding studies, we design two conditional normalizing flows to generate urban functional zones and land-use configurations in two stages which achieve great success in the urban planning task.

\vspace{-0.3cm}
\section{Preliminaries}
\vspace{-0.1cm}

\begin{figure}[t]
	\centering
    \subfigure[Target Area.]{
    \begin{minipage}[ht]{0.45\linewidth}
    \centering
    \includegraphics[width=0.8in]{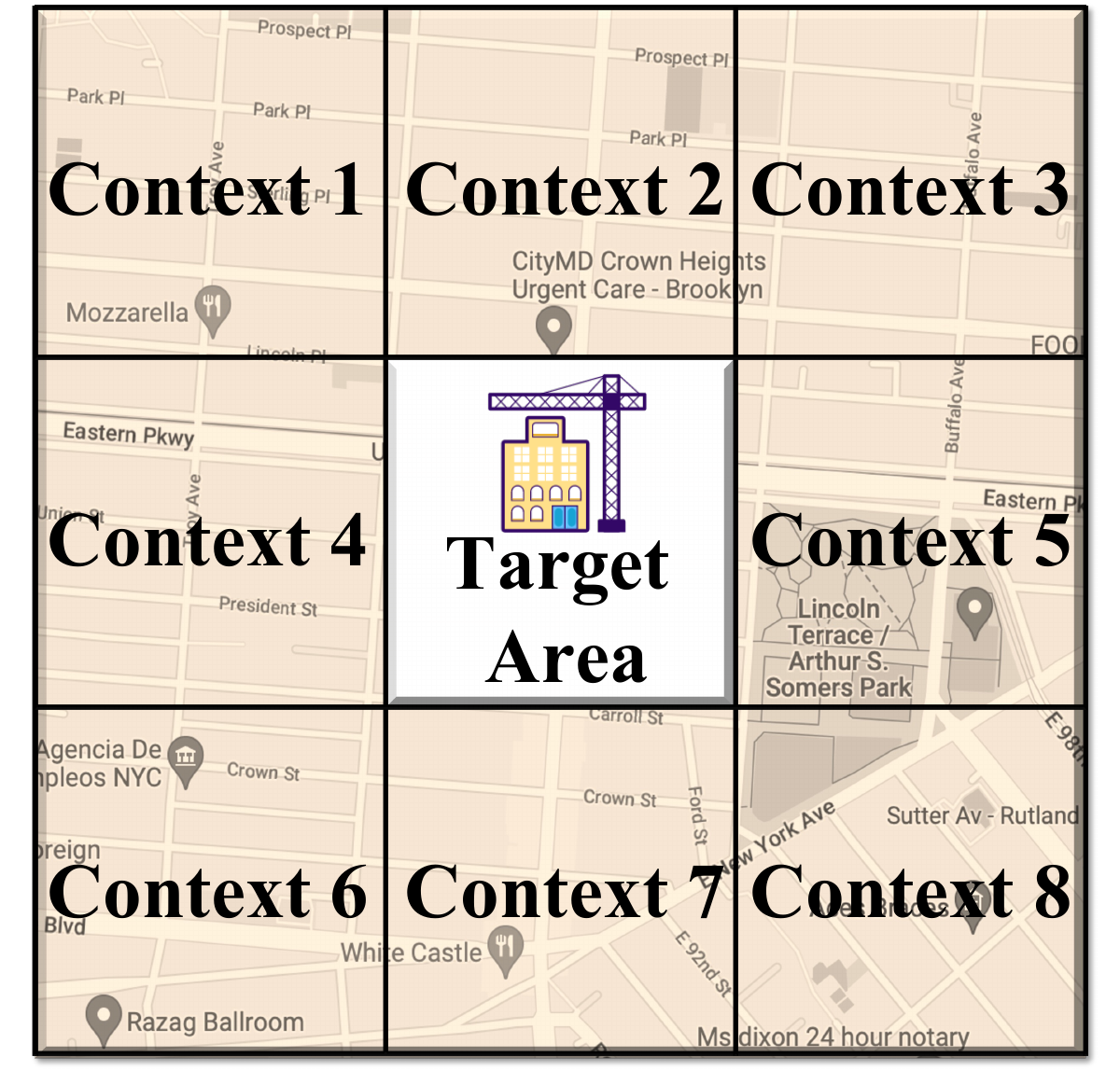}
    \vspace{1mm}
    \label{target_area}
    \end{minipage}%
    }%
    \subfigure[Surrounding Contexts.]{
    \begin{minipage}[ht]{0.45\linewidth}
    \centering
    \includegraphics[width=1.3in]{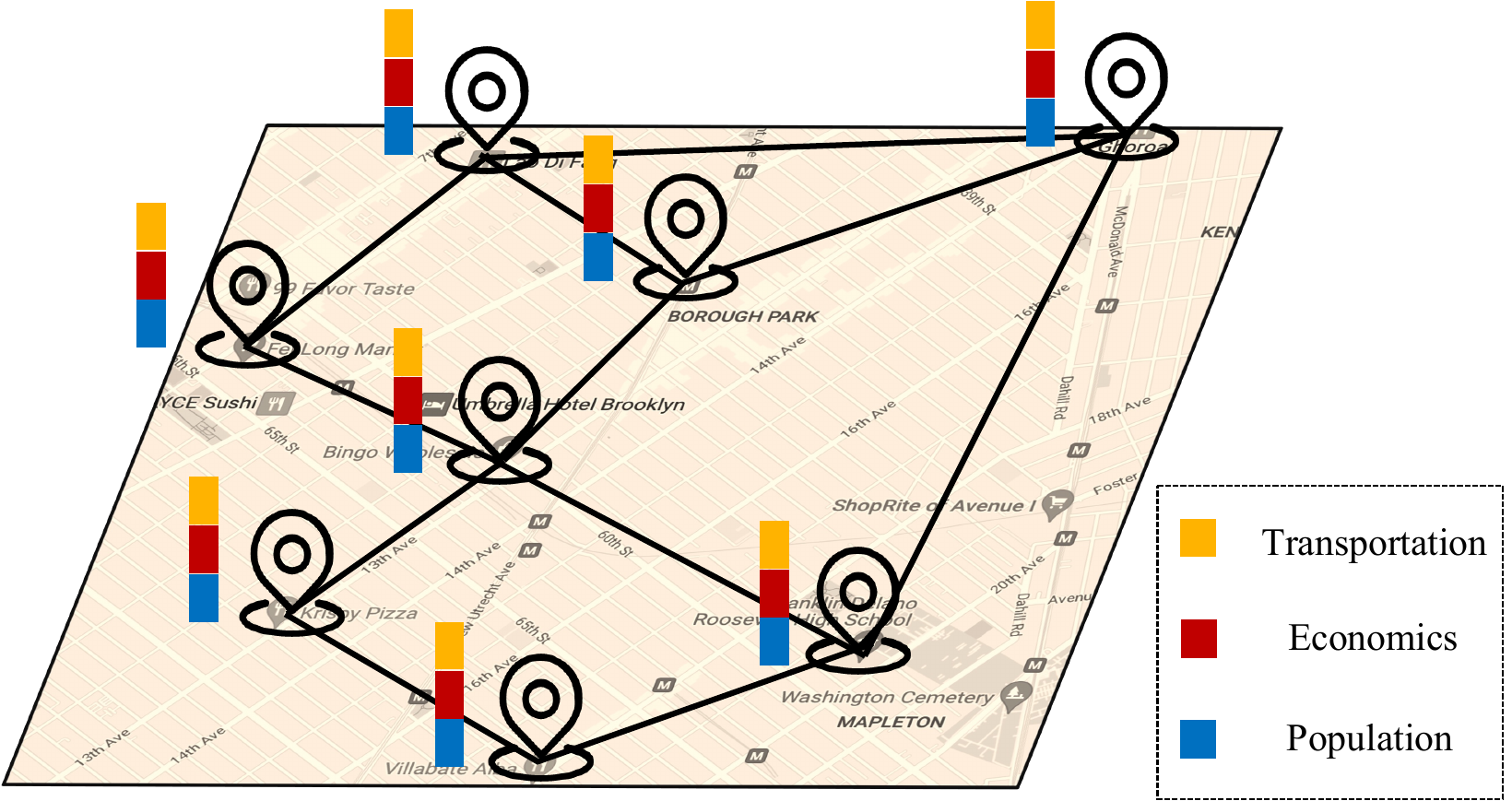}
    \label{spatial}
    \vspace{2.7mm}
    \end{minipage}%
    }%
    \\
    \subfigure[Urban Function Zone.]{
    \begin{minipage}[ht]{0.45\linewidth}
    \centering
    \includegraphics[width=1.3in]{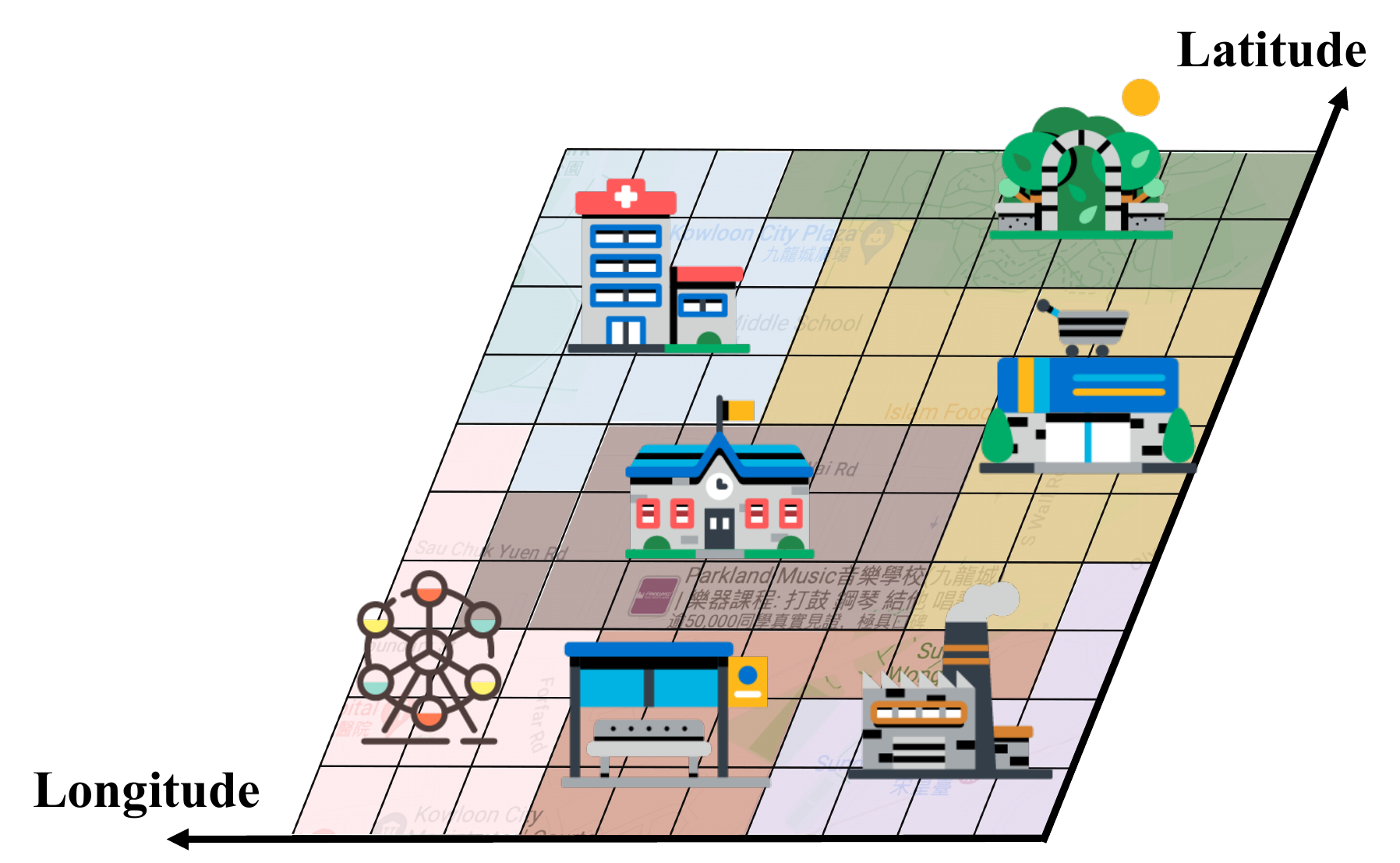}
    \label{ufz}
    \vspace{2.4mm}
    \end{minipage}%
    }
    \subfigure[Land-use Configuration.]{
    \begin{minipage}[ht]{0.45\linewidth}
    \centering
    \includegraphics[width=1.3in]{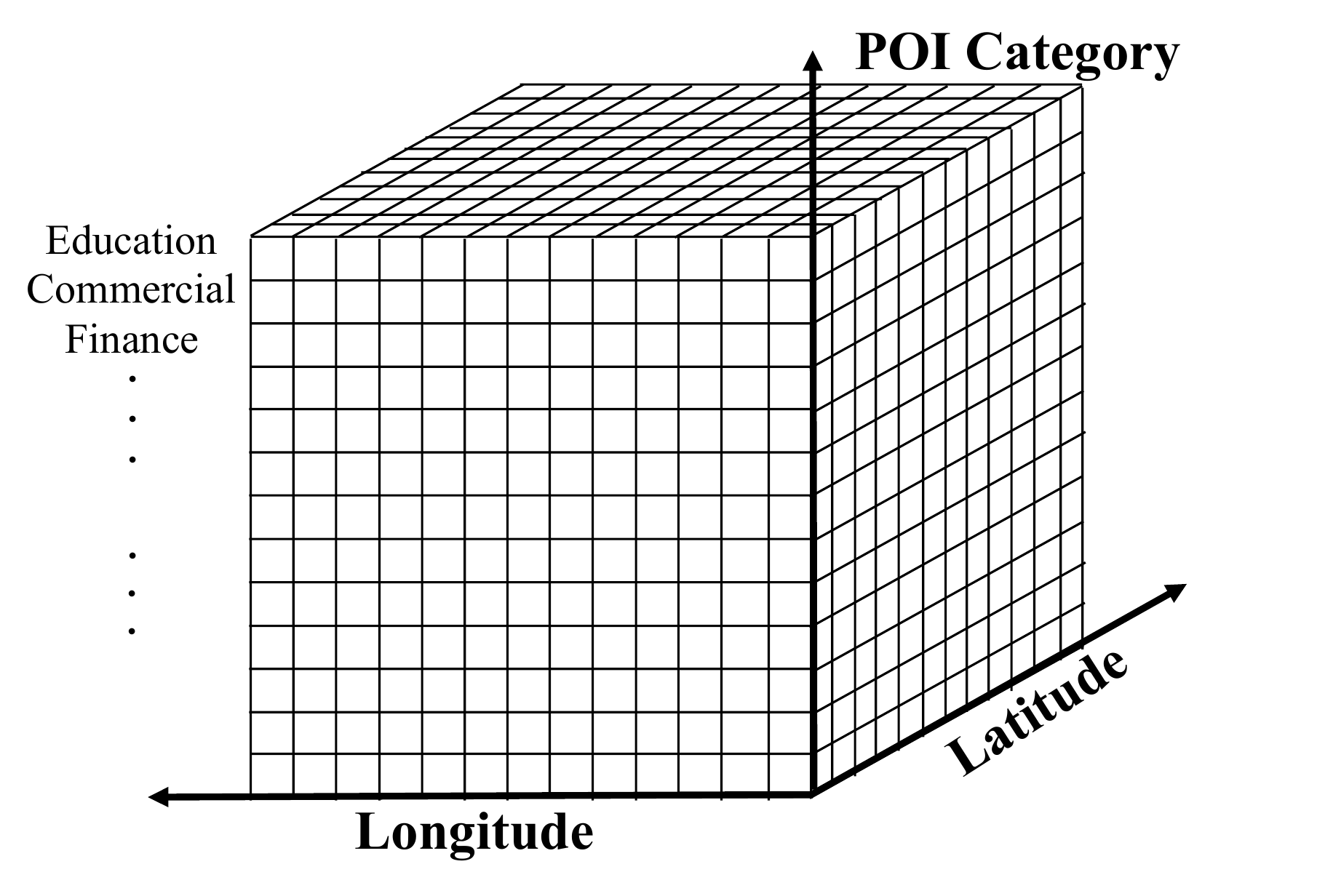}
    \label{luc}
    \end{minipage}%
    }
    \centering
    \vspace{-3mm}
    \caption{Illustration of related concepts.}
    \label{preli}
\vspace{-7mm}
\end{figure}

\paragraph{Target Area and Surrounding Contexts}
The \textit{target area} represents an empty square geographical area we aim to plan, which is encompassed by another context area named \textit{surrounding contexts} from different directions, as shown in Figure \ref{target_area}.
We store them in a spatial attributed graph illustrated by Figure \ref{spatial}, where a geographical region is represented as a node, the spatial relationship of two regions as an edge, and node attributes as the noteworthy socioeconomic features.

\paragraph{Urban Functional Zone and Land-use Configuration} The \textit{urban functional zone} (i.e., zone-level planning) contains a particular urban function with homogenous socio-economic characteristics, indicated by
Figure \ref{ufz}.
Following the setting of~\cite{wang2021deep}, we divide a geographical area into $N \times N $ squares based on latitude and longitude, and each urban functional zone $\mathbf{U} \in \mathbb{R}^{N\times N}$ contains several of them. For the \textit{land-use configuration} (i.e., configuration-level planning), it denotes the geographical distribution of Points of Interest (POI).
We store the information in a tensor $\mathbf{X} \in \mathbb{R}^{N\times N\times P}$ with dimensions of longitude, latitude, and POI category (the division of POI based on urban functions such as financial institutions and food service) shown in Figure \ref{luc}, where $P$ is the number of POI categories. 

\paragraph{Human Guidance}
In this work, \textit{human guidance} stands for human stipulations for the urban planning process such as the required distance between buildings. To empower our model to capture the semantic meaning of human guidance,  we embed these text-based stipulations and take the embeddings as parts of the conditional inputs of our urban planning generator.

\section{Methodology}

\begin{figure*}[t]
\centering
\includegraphics[width=\linewidth]{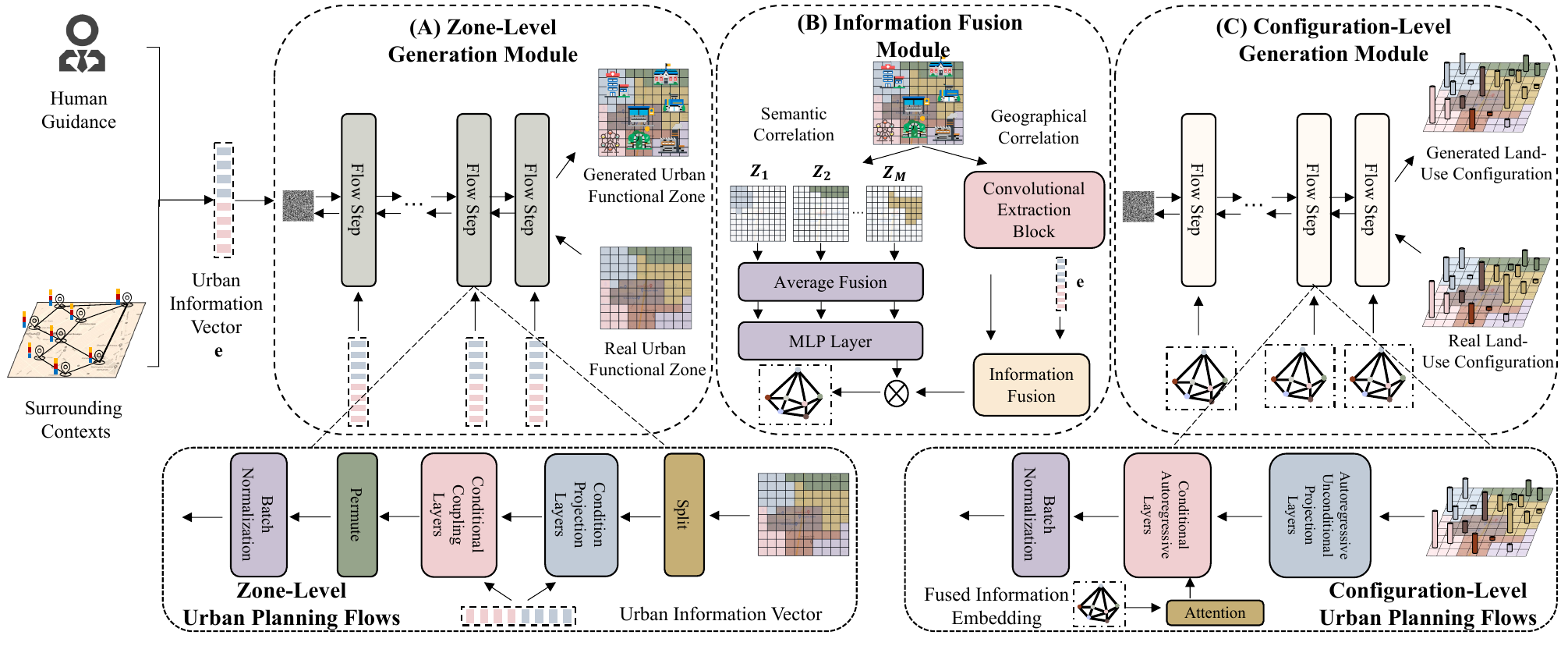}
\vspace{-9mm}
\caption{The overview of our proposed DSUF framework.}
\vspace{-4mm}
\label{framework}
\end{figure*}

Figure \ref{framework} shows the overall structure of our proposed Dual-Stage Urban Flows (DSUF) framework which mainly includes three major elements.
First, we introduce the Zone-Level Generation Module in Section \ref{sec:zone_level_generation_module}, where we propose zone-level urban planning flows to solve the data sparsity problem and obtain a rough sketch of urban planning. Then, we present the Information Fusion Module in Section \ref{information_fusion_module}, which captures the geographical information of generated urban functional zones and the semantical planning requirement.
Finally, we illustrate the Configuration-Level Generation Module in Section \ref{configuration-Level Generation Module} where we propose configuration-level urban planning flows to obtain the final land-use configurations based on the fused knowledge.

\paragraph{Problem Statement}
We intend to construct an automated urban planning generator that can produce land-use configurations based on the surrounding contexts and human guidance. In the following parts of this paper, We utilize the $i^{th}$ empty target area as an example to illustrate the generation process for convenience. Formally, we denote the features of surrounding contexts as $\mathbf{S} \in \mathbb{R}^{1\times \mathbf{D}_1}$, human guidance as $\mathbf{I} \in \mathbb{R}^{1\times \mathbf{D}_2}$, and land-use configurations as $\mathbf{X}$, where $\mathbf{D}_1$ and $\mathbf{D}_2$ are the feature dimensions of corresponding elements. Our target is to learn a mapping function $f_\theta : (\mathbf{S}, \mathbf{I}) \rightarrow \mathbf{X}$ which takes features of surrounding contexts $\mathbf{S}$ and human guidance $\mathbf{I}$ as input, and generates an output of land-use configurations $\mathbf{X}$. We embed the surrounding contexts with spatial attributed graph, and concatenate the graph embedding with the one-hot vector of human guidance, where the concatenated embedding is named as Urban Information Vector $\mathbf{e} \in \mathbb{R}^{1\times \mathbf{D}}$, and $\mathbf{D}$ is the size of feature dimension, $\mathbf{D}=\mathbf{D}_1+\mathbf{D}_2$.

\subsection{Zone-Level Generation Module} 
\label{sec:zone_level_generation_module}


Due to the sparsity of data and intricate urban planning dependencies, estimating the distribution of POIs can be a very challenging task. Instead of generating land-use configurations directly, we can imitate how humans tackle complex problems and decompose the generation process of urban planning into two more manageable sub-objectives: (1) generating a rough drawing of urban planning (i.e., zone-level planning) (2) obtaining the final urban structure (i.e., configurations-level planning) based on generated rough sketch. 
Our Zone-Level Generation Module is to accomplish the first sub-objective by modeling the conditional distribution  $p_{\mathbf{U}|\mathbf{e}}(\mathbf{U}|\mathbf{e}, \theta)$ of urban functional zones $\mathbf{U}\in \mathbb{R}^{N\times N}$ corresponding to the Urban Information Vector $\mathbf{e}$.
However, estimating such a complex distribution can be a very challenging task due to the complicated patterns of functional zones. To this end, we propose \textit{\textbf{zone-level urban planning flows}} to obtain the urban functional zone $\mathbf{U}$. Specifically,
the explicit generation process of zone-level urban planning flows $f_\theta$ can be formulated:
\begin{equation}
    p_{\mathbf{U}|\mathbf{e}}(\mathbf{U}|\mathbf{e}, \theta) = p_{\mathbf{z}}(f_{\theta}(\mathbf{U};\mathbf{e}))\left | det \frac{\partial{f_{\theta}}}{\partial{\mathbf{U}}}(\mathbf{U};\mathbf{e})\right |
\end{equation}
where $\mathbf{z}$ is the latent variable follows the Gaussian distribution,
$p_{\mathbf{U}|\mathbf{e}}(\mathbf{U}|\mathbf{e}, \theta)$ is the conditional distribution of urban functional zones.

To train the network of Zone-Level Urban Planning Flows, we minimize the negative log-likelihood function (NLL). To derive a tractable expression, we define $\mathbf{h}^0 := \mathbf{U}$, and $\mathbf{h}^K := \mathbf{z}$ and decompose the bijective mapping function $f_{\theta}$ into a set of K invertible layers, i.e. $\mathbf{h^{k+1}}=f_{\theta}^{k}(\mathbf{h}^k; \mathbf{e})$. By applying the chain rule, we can rewrite the objective function NLL as: 
\begin{equation}
\label{loss_1}
    \mathcal{L}(\theta ; \mathbf{e}, \mathbf{U})= - log p_{\mathbf{z}}(f_{\theta}(\mathbf{U};\mathbf{e})) - \sum_{k=0}^{K-1} log \left | det \frac{\partial{f_{\theta}^k}}{\partial{\mathbf{h}^{k}}}(\mathbf{h}^{k};\mathbf{e})\right|
\end{equation}

For zone-level planning, we only need to generate a rough sketch of the urban plan. Thus we construct a simple yet effective architecture for zone-level urban planning flows, inspired by \cite{dinh2014nice, dinh2016density}. Our zone-level urban planning flows are composed of conditional coupling layers, condition projection layers, and batch normalization~\cite{ioffe2015batch} (which will be introduced in Appendix \ref{batch_norm}). 

\paragraph{Conditional Coupling Layers}
We learn from the coupling layers designed by \cite{dinh2014nice, dinh2016density}. They are easily invertible and the determinant of the Jacobian matrix is tractable, which can be extended as:
\begin{equation}
\begin{aligned}
\label{ccl}
\mathbf{h}_{\mathbf{1}}^{k+1} &= \mathbf{h}_{\mathbf{1}}^{k}\\
\mathbf{h}_{\mathbf{2}}^{k+1} & = exp(f_{\theta, s}^k(\mathbf{h}_{\mathbf{1}}^{k} ;\mathbf{e}))\cdot \mathbf{h}_{\mathbf{2}}^{k} + f_{\theta, b}^k(\mathbf{h}_{\mathbf{1}}^{k} ;\mathbf{e})
\end{aligned}
\end{equation}
where $\mathbf{h}^{k}=(\mathbf{h}_{\mathbf{1}}^{k}, \mathbf{h}_{\mathbf{2}}^{k})$ is an arbitrary partition of intermediate variable in the channel dimension, and $f_{\theta, s}^k$ and $f_{\theta, b}^k$ are fully connected layers to obtain scale and bias. The determinant of Jacobian matrix can be calculated as $exp[\sum\nolimits_{j}f_{\theta, s}^k(\mathbf{h}_{\mathbf{1}}^{k} ;\mathbf{e})_j]$. Besides, we include the inverse process of Equation (\ref{ccl}) in Appendix \ref{inverse_tran}.

\paragraph{Condition Projection Layers}
One of the biggest challenges of automated urban planning is data sparsity which may impede the generation of DSUF. To alleviate this effect, we include another condition projection layer to directly capture the inherent knowledge from Urban Information Vector. 
The first partition $h_1^k$ of condition projection layers is same as Equation \ref{ccl}; the second partition is:
\begin{equation}
\begin{aligned}
\label{cpl}
\mathbf{h}_{\mathbf{2}}^{k+1} &= exp(f_{\theta, s}^k(\mathbf{e}))\cdot \mathbf{h}_{\mathbf{2}}^{k} + f_{\theta, b}^k(\mathbf{e})
\end{aligned}
\end{equation}
Here, $f_{\theta, s}^k$ and $f_{\theta, b}^k$ are fully connected layers. The calculation process of determinant of Jacobian matrix is $exp[\sum\nolimits_{j}f_{\theta, s}^k(\mathbf{e})_j]$. The inverse transformation of Condition Projection Layers is referred in Appendix \ref{inverse_tran}.
The zone-level urban planning flows are firstly optimized to generate the urban functional zones. Then, the generated results are taken for the planning process in the second stage where zone-level flows are fine-tuned with configuration-level flows for final urban planning.

\vspace{-3mm}
\subsection{Information Fusion Module} 
\vspace{-1mm}
\label{information_fusion_module}
Given the generated urban functional zones, we need to further extract the hidden planning dependencies to fuse geographical information (i.e., locations of urban functional zones) and semantic correlation (i.e., the urban functionalities of zone-level planning).
For this aim, we utilize a convolution extraction block for geographical information and a semantic projection block for abstracting semantic correlation, respectively.

\paragraph{Convolution Extraction Block}
The geographical locations of each functional zone are also crucial for the planning process. For instance, the rent price of land around commercial zones is generally higher than in other parts of cities. Hence, industrial buildings normally are located far away from the city center to reduce the rent cost, and commercial buildings are usually planned in the city center to attract more consumers. For this aim, we propose a \textit{\textbf{convolution extraction block}} to access the geographical locations and capture their geographical correlations for representation. Specifically, we first employ one layer of 2D convolution~\cite{krizhevsky2017imagenet} and layer normalization~\cite{ba2016layer} to process the input; then we adopt $N$ ConvNeXt~\cite{liu2022convnet} layers and $N-1$ down-sample layers~\cite{liu2022convnet} for correlation extraction. The $\ell$-th ConvNext layer is written as:  
\begin{equation}
\begin{split}
\mathbf{s}^{\ell} &= \rm{GELU}(\rm{Conv2d}(\rm{LayerNorm
}\\&(\rm{DepthwiseConv2d}(\mathbf{o}^{\ell-1})))), \\ \mathbf{o}^{\ell} &= \rm{DropPath
}(\rm{LayerScale}(\rm{Conv2d}(\mathbf{s}^{\ell})))+ \mathbf{o}^{\ell-1}
\end{split}
\end{equation}
where $\mathbf{o}^{\ell}$ is the output of $\ell$-th ConvNext layer, $\mathbf{s}^{\ell}$ is hidden state, GELU is the activation function, $\rm{DepthwiseConv2d}$ is the depthwise spatial convolution, $\rm{DropPath}$ randomly drops samples with a certain probability, $\rm{LayerScale}$ adds a learnable matrix on output to improve training dynamics. For the first $N-1$ layers, each ConvNext layer is followed by a down-sample layer. And the $\ell$-th down-sample layer can be formulated as: $
    \mathbf{o}^{\ell} = \rm{Conv2d}(\rm{LayerNorm}(\mathbf{o}^{\ell}))$.
In the final layer, we obtain the geographical embedding $\mathbf{o}$ by: 
    $\mathbf{o} = \rm{Linear}(\rm{LayerNorm}(\rm{GlobalAvgPooling}(\mathbf{o}^N)))$.
where $\mathbf{o}^N$ is the output of $N$-th ConvNext layer. $\rm{GlobalAvgPooling}$ represents the global average pooling operation. And $\mathbf{o}\in \mathbb{R}^{1\times D}$.

\paragraph{Semantic Projection Block}
To design a feasible urban plan, human experts usually take urban functionalities such as low criminal rates into account. 
This motivates us to propose the semantic projection block to abstract the semantic dependencies among different functional zones. According to the urban function zones generated in the first stage, we divide the zone-level planning into $M$ subareas based on their functionalities, $\mathbf{Z} = [\mathbf{Z}_1, \mathbf{Z}_2,..., \mathbf{Z}_M]$, where $\mathbf{Z}\in \mathbb{R}^{M\times N \times N}$. 
In addition, to combine the Urban information Vector $\mathbf{e}$ and geographical embedding $\mathbf{o}$, we use two learnable parameters to reflect their weights, i.e., $\mathbf{W}_s$ and $\mathbf{W}_g$. 
The process of semantic projection is:
\begin{equation}
    \mathbf{c} = \rm{softmax}(\rm{avg}(\mathbf{Z})\cdot \mathbf{W}_z)\cdot (\mathbf{W}_s \cdot \mathbf{e}+\mathbf{W}_g \cdot \mathbf{o})
\end{equation}

Here, $\rm{avg}(\cdot)$ column-wisely averages the contents of functional zones, and reshapes $\mathbf{Z}$ as $\mathbb{R}^{M\times N}$. $\mathbf{W}_z\in \mathbb{R}^{N\times 1}$ is the weight matrix. And $\mathbf{c}\in \mathbb{R}^{M\times D}$.
Ultimately, we can obtain the fused information embedding $\mathbf{c}$ with semantic projects and geographical embedding. 

\vspace{-3mm}
\subsection{Configuration-Level Generation Module}
\vspace{-1mm}
\label{configuration-Level Generation Module}

In the Zone-Level Generation Module, we obtain the rough drawings of our construction plans. However, they are lacking of details and need to be further processed to reveal more meticulous configurations. Thus, we design the configuration-level generation module for configuration-level planning $\mathbf{X}\in \mathbb{R}^{N\times N \times P}$ ($P$ is the number of POI categories). For the consistency of two stages in generation, we propose another \textit{\textbf{configuration-level urban planning flows}} to model the intricate conditional distribution of land-use configurations $p_{\mathbf{X}|\mathbf{c}}(\mathbf{X}|\mathbf{c}, \theta^{\prime})$ conditional on the fusion information embedding $\mathbf{c}$. We can write the generation procedure of configuration-level urban planning flows as $p_{\mathbf{X}|\mathbf{c}}(\mathbf{X}|\mathbf{c}, \theta^{\prime}) = p_{z^{\prime}}(g_{{\theta}^\prime}(\mathbf{X};\mathbf{c}))\left|det\frac{\partial g_{\theta^{\prime}}}{\partial \mathbf{X}}(\mathbf{X};\mathbf{c})\right|$.
And $g_{\theta^{\prime}}$ stands for the bijective mapping function and $\mathbf{z}^{\prime} = g_{\theta^{\prime}}(\mathbf{X})$. For optimization, we minimize the negative log-likelihood function, the objective function can be formulated as:
$
    \mathcal{L}(\theta^{\prime} ; \mathbf{c}, \mathbf{X}) = - log p_{\mathbf{z}^{\prime}}(g_{\theta^{\prime}}(\mathbf{X};\mathbf{c})) -  \sum_{k=0}^{K^{\prime}-1}log \left | det \frac{\partial{g_{\theta^{\prime}}^k}}{\partial{\mathbf{m}^k}}(\mathbf{m}^k;\mathbf{c})\right |$.
Here, we decompose $g_{\theta^{\prime}}$ as $g_{\theta^{\prime}} = g_{\theta^{\prime}}^0 \circ g_{\theta^{\prime}}^1 \circ \cdots \circ g_{\theta^{\prime}}^{K^{\prime}}$ and define $m^0 := \mathbf{X}$ and $m^{K^{\prime}} := \mathbf{z}^{\prime}$. 
We update the parameters of zone-level urban planning flows to avoid error accumulation when we are training the configuration-level urban planning flows. Our configuration-level urban planning flows consist of conditional autoregressive layers, autoregressive condition projection layers and batch normalization.

\paragraph{Conditional Autoregressive Layers}
The fused information embedding $\mathbf{c}$ generated in the previous Information Fusion Module may also suffer from information redundancy. To tackle it, we introduce the multi-attentions proposed by ~\cite{vaswani2017attention} to extract the most essential information. We acquire the attention score matrix $\mathbf{A}\in \mathbb{R}^{M\times D}$ by $\mathbf{A} = \rm{softmax}(\frac{\mathbf{Q}\cdot{\mathbf{K}^{T}}}{\sqrt{d_k}})\cdot \mathbf{V}$, where $\mathbf{Q}$, $\mathbf{K}$, $\mathbf{V}$ represents the query, key, and the value matrix respectively, and all of them come from $\mathbf{c}$ by linear transformation.
Then, we intend to obtain a detailed urban plan for target areas accurately conditional on the above attention matrix. 
Inspired by MAF~(\cite{papamakarios2017masked}), we design our conditional autoregressive layers and rewrite the conditional probability as $p_{\mathbf{X}|\mathbf{A}}(\mathbf{X}|\mathbf{A}, \theta^{\prime})= \prod\nolimits_i p_{\mathbf{X_i}|\mathbf{A}}(\mathbf{X}_i|\mathbf{X}_{1:i-1}, \mathbf{A}, \theta^{\prime})$. By using the following recursion, we generate the land-use configurations: 
\begin{equation}
\label{cal_forward}
    \mathbf{m}_i^{k+1} = \mathbf{m}_i^{k} \cdot exp(g_{\theta^{\prime}, s_i}^k(\mathbf{m}_{1:i-1}^{k};\mathbf{A})) + g_{\theta^{\prime}, b_i}^k(\mathbf{m}_{1:i-1}^{k};\mathbf{A})
\end{equation}
where, $g_{\theta^{\prime}, s_i}^k$ and $g_{\theta^{\prime}, b_i}^k$ are unconstrained functions and are implemented with masking following the approach of MADE~\cite{germain2015made}. The log determinant can be calculated as $log \left | det \frac{\partial{g_{\theta^{\prime}}^k}}{\partial{\mathbf{m}^k}}(\mathbf{m}^k;\mathbf{c})\right | = -\sum \nolimits_i g_{\theta^{\prime}, s_i}(\mathbf{m}_{1:i-1}^k;\mathbf{A})$.

\paragraph{Autoregressive Unconditional Projection Layers}
Due to the sparsity of land-use configurations, the tensor that stores the information is a sparse matrix. To better access the information directly, we design autoregressive unconditional projection Layers. Our configuration-level urban planning flows can better capture the planning dependencies with them.
\begin{equation}
\label{aupl_forward}
    \mathbf{m}_i^{k+1} = \mathbf{m}_i^{k} \cdot exp(g_{\theta^{\prime}, s_i}^k(\mathbf{m}_{1:i-1}^k)) + g_{\theta^{\prime}, b_i}^k(\mathbf{m}_{1:i-1}^k)
\end{equation}
where, the design of $g_{\theta^{\prime}, s_i}^k$ and $g_{\theta^{\prime}, b_i}^k$ is similar to conditional autoregressive layers. The log determinant is formulated as $-\sum \nolimits_i g_{\theta^{\prime}, s_i}(\mathbf{m}_{1:i-1}^k)$.

We also employ Batch Normalization for configuration-level urban planning flows, the calculation is similar to zone-level urban planning flows.

\section{Experiments}
\begin{table}[t]
\centering
\scriptsize
\setlength{\tabcolsep}{1pt}
\caption{Performance Compared to CLUVAE} 
\label{table_main}
\begin{tabular}{c c c c c}

\toprule[1.5pt]
Model name & AVG\_HD & \makecell{AVG\_WD\\($1\times10^{-5}$)} & \makecell{Improvement of \\AVG\_HD} & \makecell{Improvement of \\AVG\_WD}\\
\midrule
   DSUF& 1.81& 2.41& -43.79\%&-22.51\%\\
   LUCGAN&6.87&3.27&+113.35\%&+5.14\%\\
   CVAE&8.10&4.12&+151.55\%&+32.48\%\\
   CGAN&7.41&4.23&+130.12\%&+36.01\%\\
   DCGAN&5.78&3.56&+79.50\%&+14.47\%\\
   WGAN&4.89&2.89&+51.86\%&-7.07\%\\
   CLUVAE&3.22&3.11&--&--\\
   
\bottomrule[1.5pt]
\vspace{-8mm}
\end{tabular}
\end{table}
\paragraph{Data Description}
We collect data from Beijing by the following process: We first crawl 2990 residential communities from soufun.com and collect 328,668 POIs as land-use configurations with their latitudes, longitudes and POI categories from openstreetmap.org ~\cite{wang2020reimagining}. In Appendix \ref{data_des}, we demonstrate the 20 different POI categories we collect. To construct urban functional zones, we download taxi trajectories from the T-drive project~\cite{yuan2010t} as well as road networks and POIs from openstreetmap.org~\cite{yuan2014discovering}. The \textit{green level rate}, which measures the amount of greenery in the range of $[0\sim1]$, is an essential human stipulation for urban planning. So we employ the green rate level related to the obtained residential communities as our human guidance. Ultimately, we utilize taxi trajectories, POIs, mobile checkins crawled from weibo.com, and housing price data crawled from soufun.com to obtain the socioeconomic features as surrounding contexts.

\paragraph{Evaluation Metrics}
In these experiments, we consider the label of these green rate levels as human guidance. We divide the green level rate of our dataset into five green rate levels: Green0, Green1, Green2, Green3, Green4. The green level rate of land-use configuration increases monotonically with the green rate level index. To evaluate the performance of our generative models, we calculate the distribution distance between generated land-use configurations and ground truths under the setting of different green rate levels. 
Specifically, we utilize three evaluation metrics in our work: 1) Average Kullback-Leibler (AVG$\_$KL) Divergence~\cite{kullback1951information}.
2) Average Hellinger Distance (AVG$\_$HD)~\cite{hellinger1909neue}.
3) Average Wasserstein Distance (AVG$\_$WD)~\cite{vaserstein1969markov}.
More evaluation details are revealed in Appendix~\ref{Eva_M}.

\paragraph{Baseline Models}
To evaluate the effectiveness of our proposed DSUF, we include the following widely used deep learning generative models as baselines: \noindent{\bf (1) LUCGAN~\cite{wang2020reimagining}}, can generate land-use configurations for a target area based on given socioeconomic features. 
\noindent{\bf (2) CLUVAE~\cite{wang2021deep}}utilizes a VAE-based model to obtain urban plans according to socioeconomic features as well as human guidance. 
\noindent{\bf (3) CVAE~\cite{sohn2015learning}}
can generate excellent samples conditional on the inputs via VAE. 
\noindent{\bf (4)CGAN~\cite{mirza2014conditional}}  
is the conditional version of GAN to access conditional results. 
\noindent{\bf (5) DCGAN~\cite{radford2015unsupervised}} 
can capture the geospatial pattern by employing the convolutional mechanism. 
\noindent{\bf (6) WGAN~\cite{arjovsky2017wasserstein}} 
is designed to be more stable compared to traditional GAN by optimizing Wasserstein loss.

\subsection{Compared Experiments}
\begin{figure}[t]
	\centering
    \subfigure[AVG\_KL.]{
    \begin{minipage}[ht]{0.3\linewidth}
    \centering
    \includegraphics[width=1.1in]{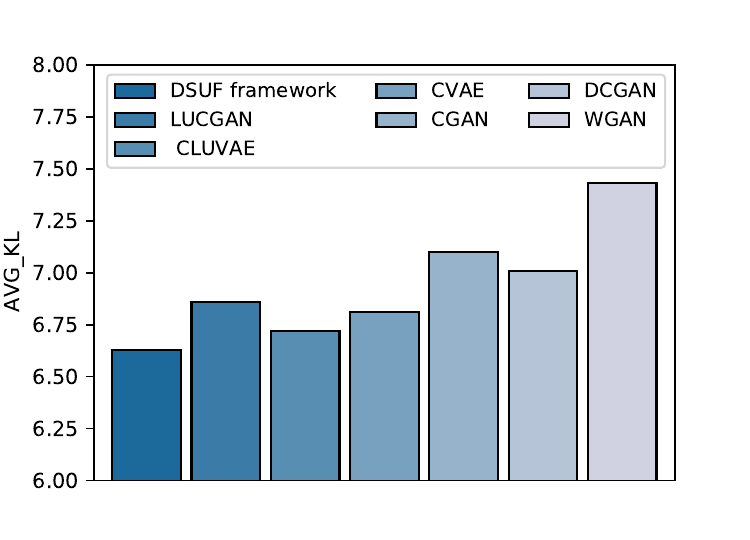}
    \end{minipage}%
    }%
    \subfigure[AVG\_HD.]{
    \begin{minipage}[ht]{0.3\linewidth}
    \centering
    \includegraphics[width=1.1in]{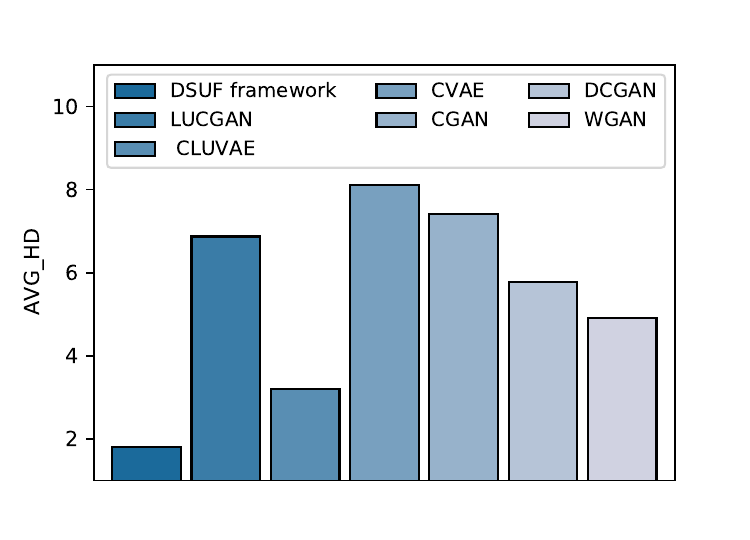}
    \end{minipage}%
    }%
    \vspace{-1mm}
    \subfigure[AVG\_WD]{
    \begin{minipage}[ht]{0.3\linewidth}
    \centering
    \includegraphics[width=1.1in]{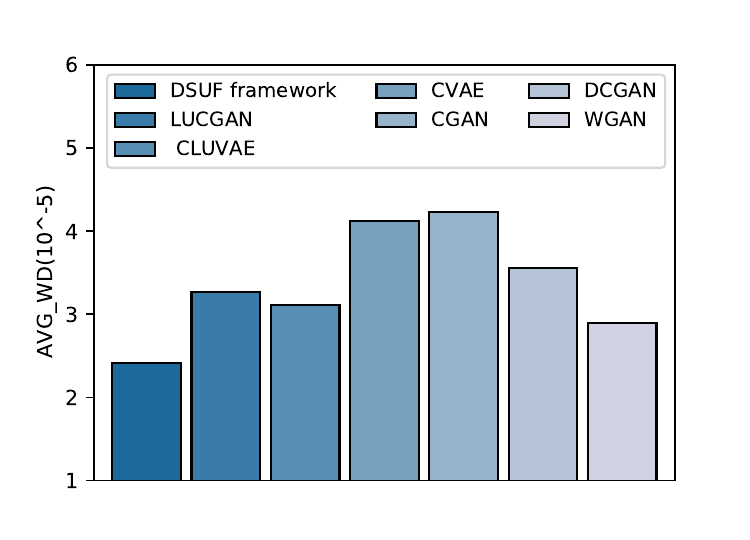}
    \end{minipage}%
    }
    \vspace{-3mm}
    \centering
    \caption{Overall Performance of DSUP Flows and other baselines in terms of all evaluation metrics.}
    \label{main}
\vspace{-4.5mm}
\end{figure}


   

\paragraph{On Effectiveness Against Baseline Methods.}
The final results of our proposed DSUF framework are reported in Figure \ref{main}. In general, It is easy to observe that the DSUF framework can outperform all the baselines at convincing levels. Especially, compared to CLUVAE, average Hellinger Distance decrease 43.79\% (3.22 $\Rightarrow$ 1.81), average Wasserstein Distance decrease 22.51 \% (3.11$\times 10^{-5}$ $\Rightarrow$ 2.41$\times 10^{-5}$) as shown in Table~\ref{table_main}. This observation demonstrates that with a dual-stage generation process and Information Fusion Module, our proposed DSUF framework can effectively capture the hierarchical relationship between urban functional zones and land-use configurations, and model the planning dependencies among diverse urban functional zones. By taking these correlations into account, the DSUF framework can generate high-quality land-use configurations.

\paragraph{On Robustness Against Different Grid Sizes and Data Partitions.}
\begin{figure}[t]
	\centering
    \subfigure[AVG\_KL.]{
    \begin{minipage}[ht]{0.3\linewidth}
    \centering
    \includegraphics[width=1.1in]{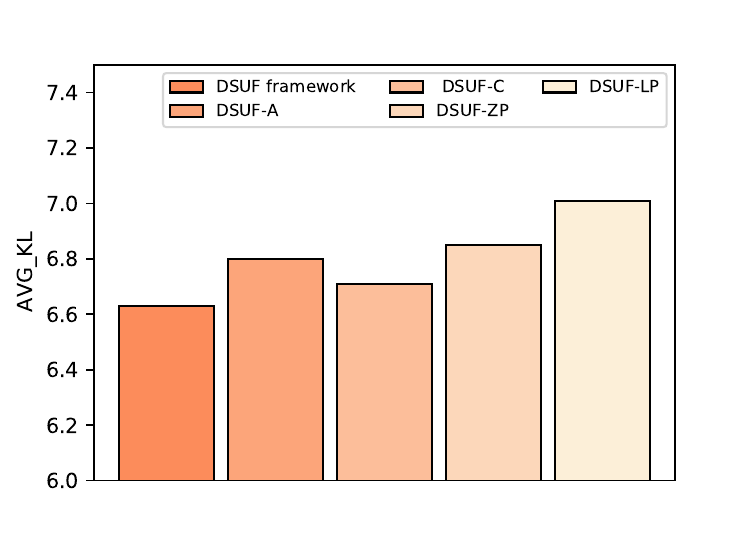}
    \end{minipage}%
    }%
    \subfigure[AVG\_HD.]{
    \begin{minipage}[ht]{0.3\linewidth}
    \centering
    \includegraphics[width=1.1in]{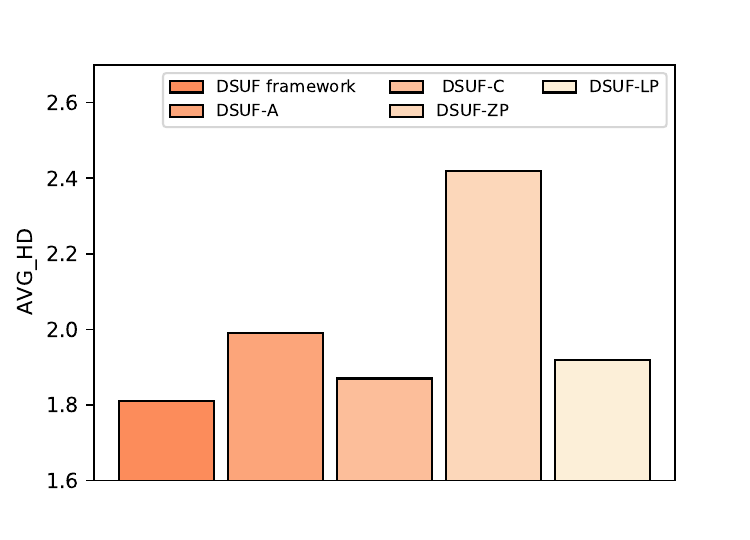}
    \end{minipage}%
    }%
    \subfigure[AVG\_WD.]{
    \begin{minipage}[ht]{0.3\linewidth}
    \centering
    \includegraphics[width=1.1in]{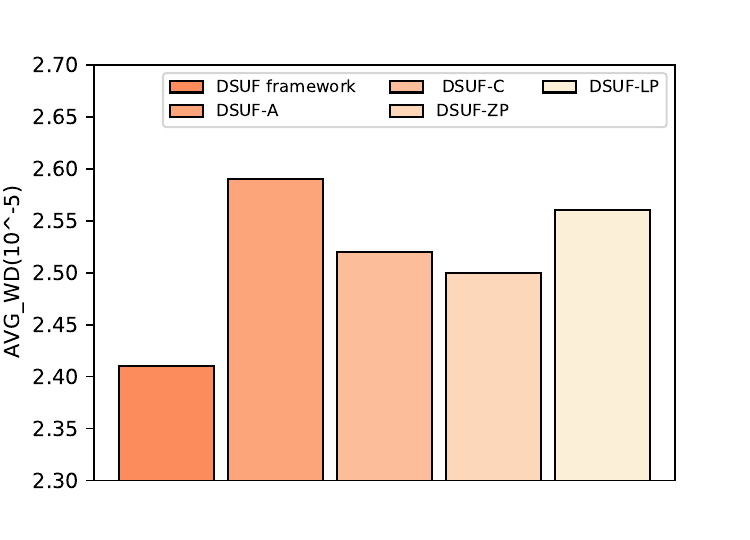}
    \end{minipage}%
    }
    \vspace{-3mm}
    \centering
    \caption{Ablation Study for DSUP Flows in terms of all evaluation metrics.}
    \label{abl}
\vspace{-3mm}
\end{figure}
\begin{figure}[t]
	\centering
    \subfigure[AVG\_KL.]{
    \begin{minipage}[ht]{0.3\linewidth}
    \centering
    \includegraphics[width=1.1in]{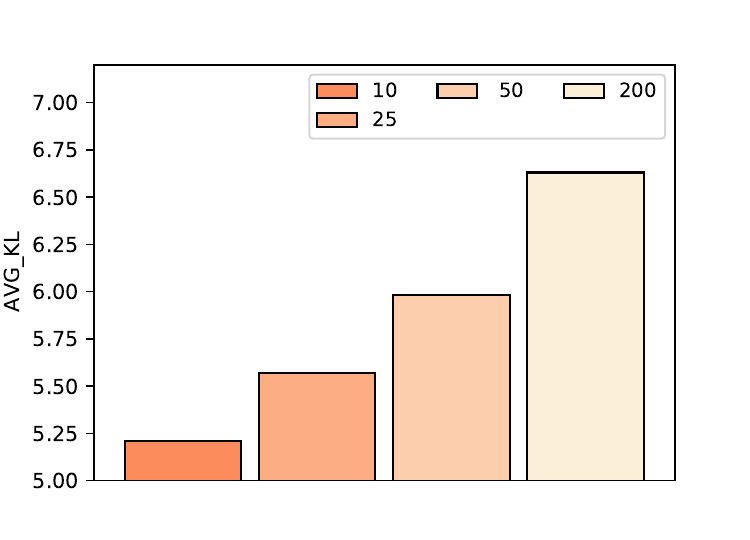}
    \end{minipage}%
    }%
    \subfigure[AVG\_HD.]{
    \begin{minipage}[ht]{0.3\linewidth}
    \centering
    \includegraphics[width=1.1in]{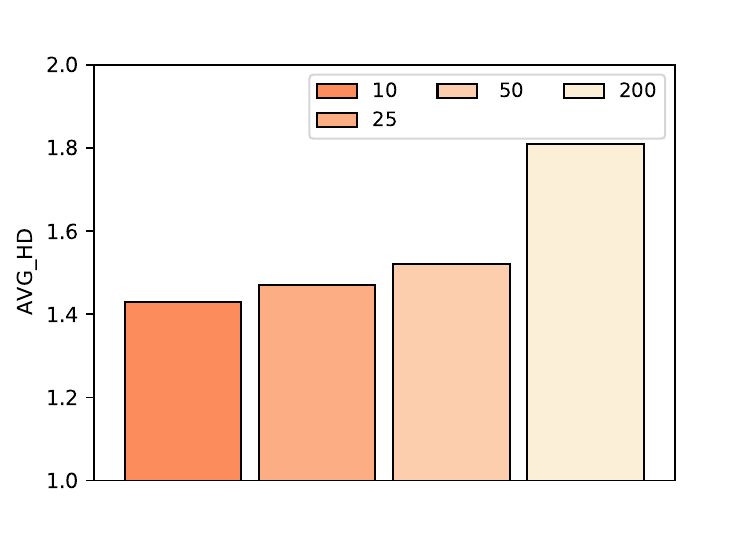}
    \end{minipage}%
    }%
    \subfigure[AVG\_WD.]{
    \begin{minipage}[ht]{0.3\linewidth}
    \centering
    \includegraphics[width=1.1in]{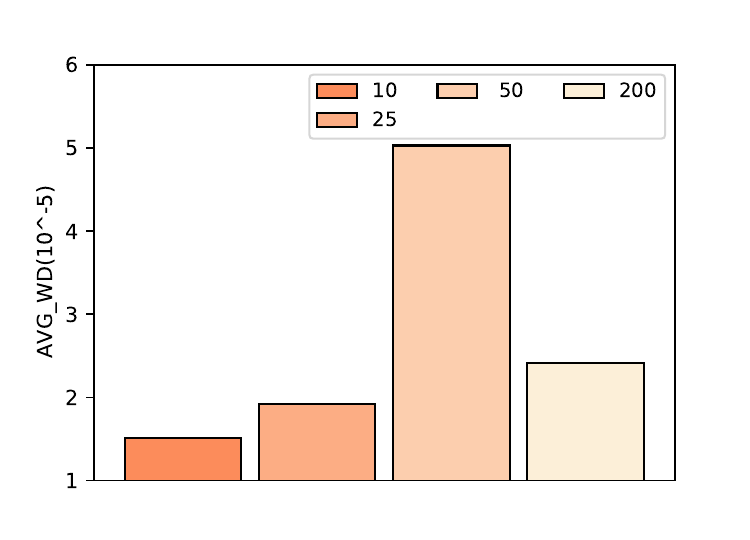}
    \end{minipage}%
    }
    \vspace{-3mm}
    \centering
    \caption{Robustness check for different square sizes of land-use configurations.}
    \label{ro}
\vspace{-7mm}
\end{figure}
We design the robustness check for our approach in this part by changing the value of $N$ which is used as the partition number of the geographical areas. In our setting, the larger partition number $N$ refers to subtler land-use configurations. $N$ is set to be 100 for other experiments, while in this experiment, we change the value from 5 to 100. The result is indicated by Figure \ref{ro}. This result demonstrates that the DSUF framework can effectively capture the planning dependencies and stably generate urban solutions. One possible reason for the stable generation process is that flow-based models only employ one loss which is easy to train, and the bijective mapping between real space and latent space can guarantee the stability of generation.  

\paragraph{On Traceability: Our Method Can Enhance the Traceability and Visualization of Generation Flow.}

\begin{figure}[t]
    \centering
    \subfigure[AVG\_KL.]{\includegraphics[width=0.155\textwidth]{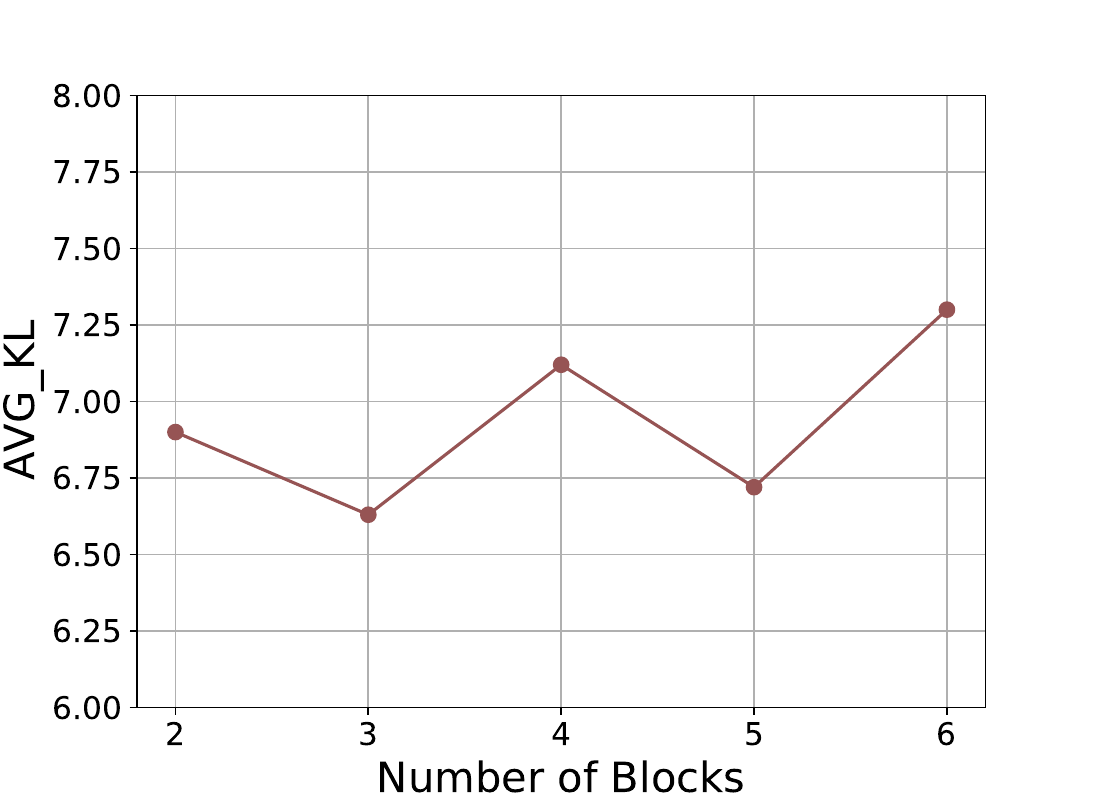}}
    \subfigure[AVG\_HD.]{\includegraphics[width=0.155\textwidth]{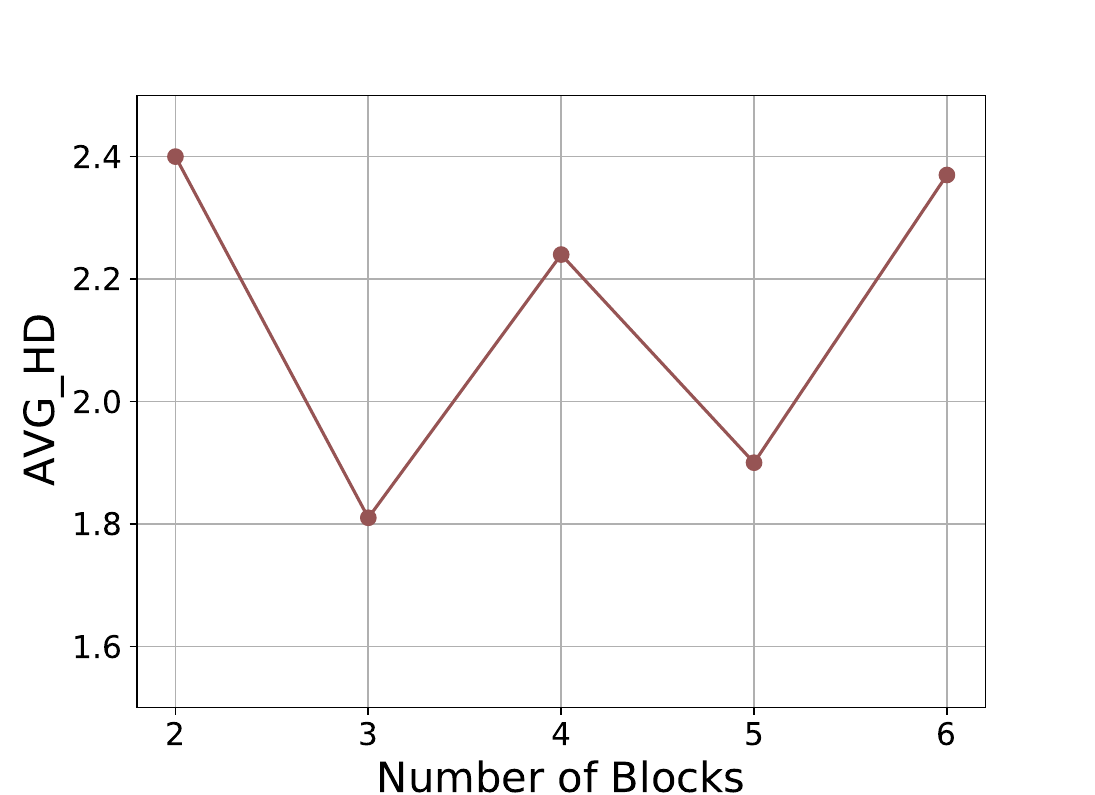}}
    \subfigure[AVG\_WD.]{\includegraphics[width=0.155\textwidth]{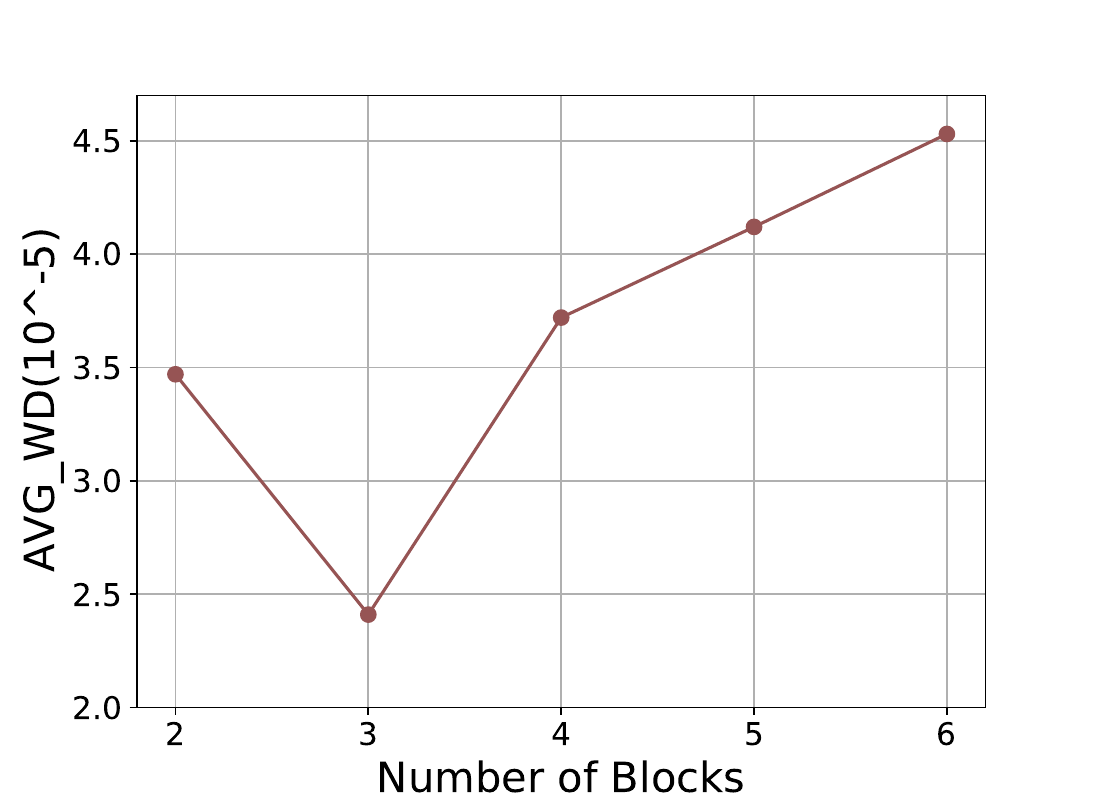}}
    \vspace{-3mm}
    \caption{Sensitivity test of number of blocks in configuration-level urban planning flows.}
    \label{sen}
    \vspace{-5mm}
\end{figure}

\begin{figure}[t]
\centering
\includegraphics[width=3.0in]{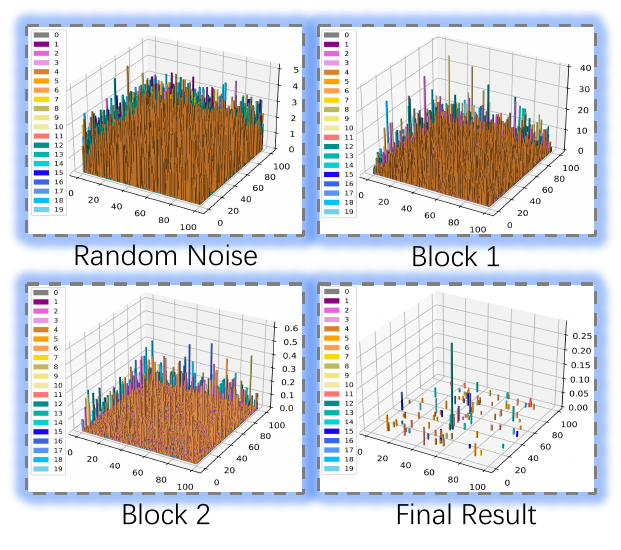}
\vspace{-6mm}
\caption{The visualization of the generation process. We illustrate how configuration-level urban planning flows generate land-use configurations.}
\label{generation process}
\vspace{-6.5mm}
\end{figure}

As mentioned previously, normalizing flows is a set of bijective functions. We can investigate the calculation result of each step on account of the flow structure. Our proposed configuration-level urban planning flows can generate urban solutions step by step, which guarantees the traceability of the generation process.
This qualitative study aims to illustrate the generation process of our proposed configuration-level urban planning flows. Figure \ref{generation process} exhibits the visualizations of the generation process from random noise (the leftmost subfigure) to desirable land-use configurations (the rightmost subfigure). The color legend in each subfigure represents the corresponding relationship between colors and POI categories, and the height of each bar is the number of POI categories. We can observe that configuration-level urban planning flows can adjust the proportion of each category of POI in the target area step by step and ultimately generate the desirable urban solution. The visible generation process can increase the reliability of the DSUF framework.
\vspace{-3mm}
\subsection{Ablation Study and Additional Analysis}
\vspace{-1mm}

\paragraph{Examining the Effectiveness of different components.} 
\begin{figure}[t]
    \centering
    \subfigure[AVG\_KL.]{\includegraphics[width=0.155\textwidth]{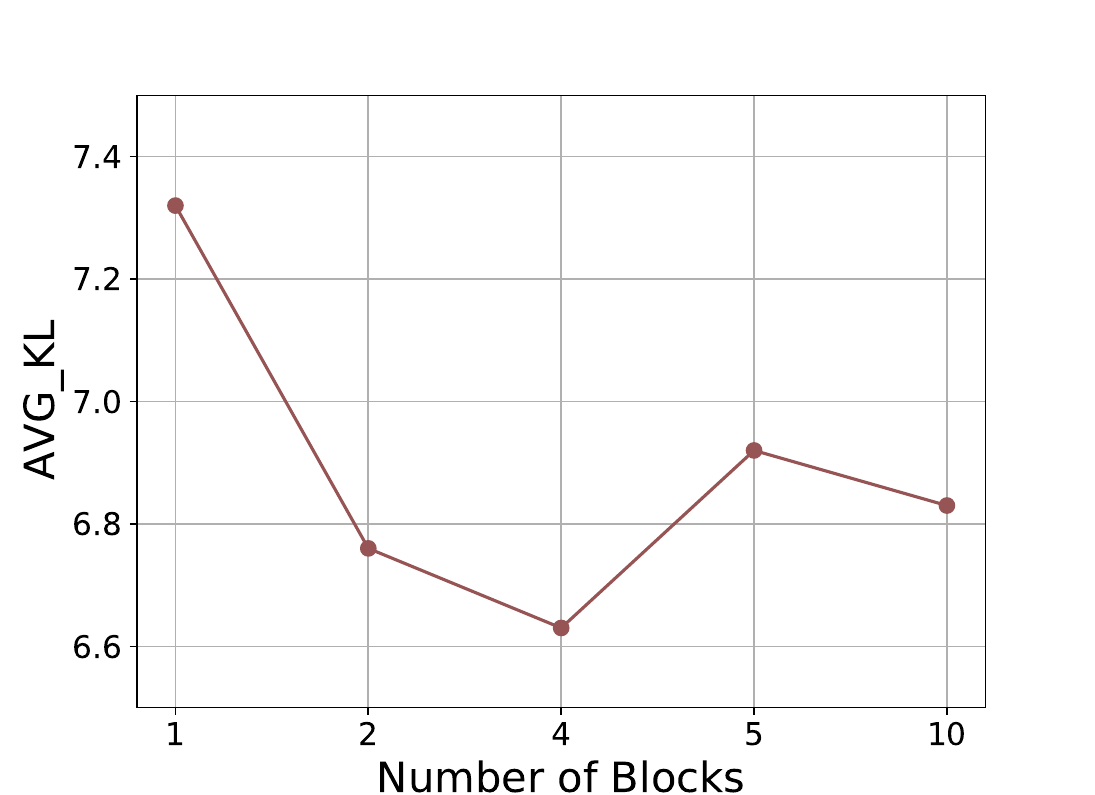}}
    \subfigure[AVG\_HD.]{\includegraphics[width=0.155\textwidth]{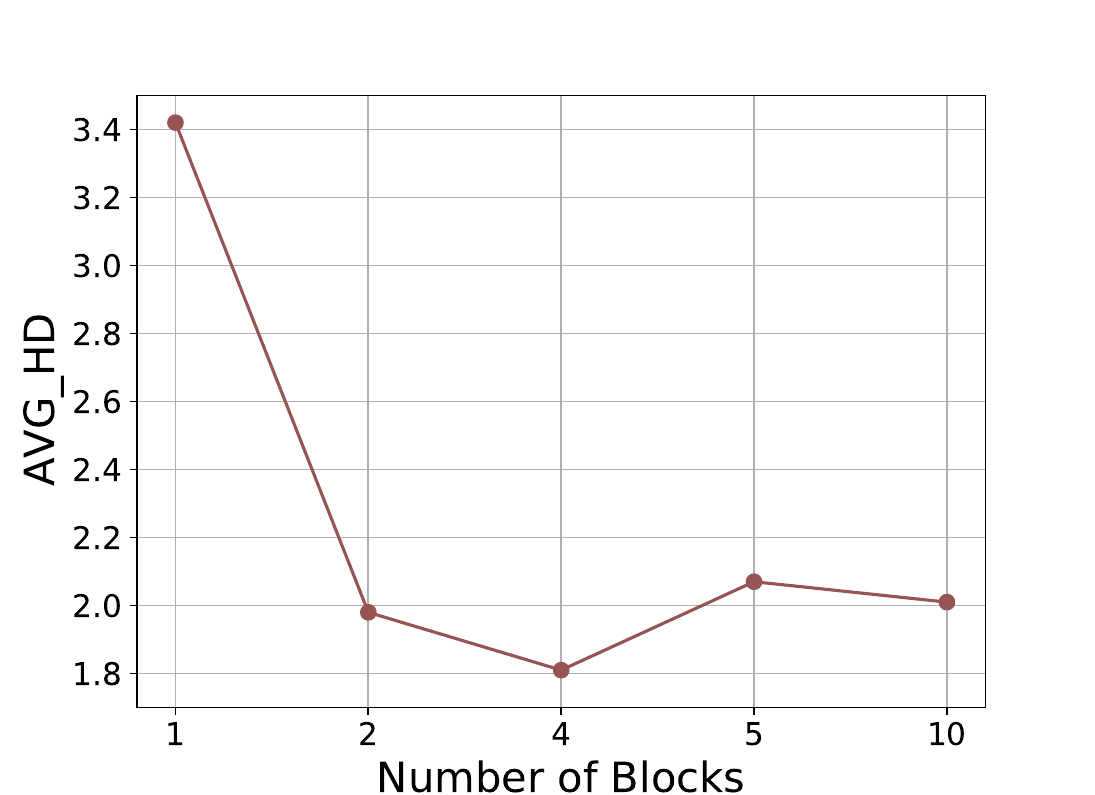}}
    \subfigure[AVG\_WD.]{\includegraphics[width=0.155\textwidth]{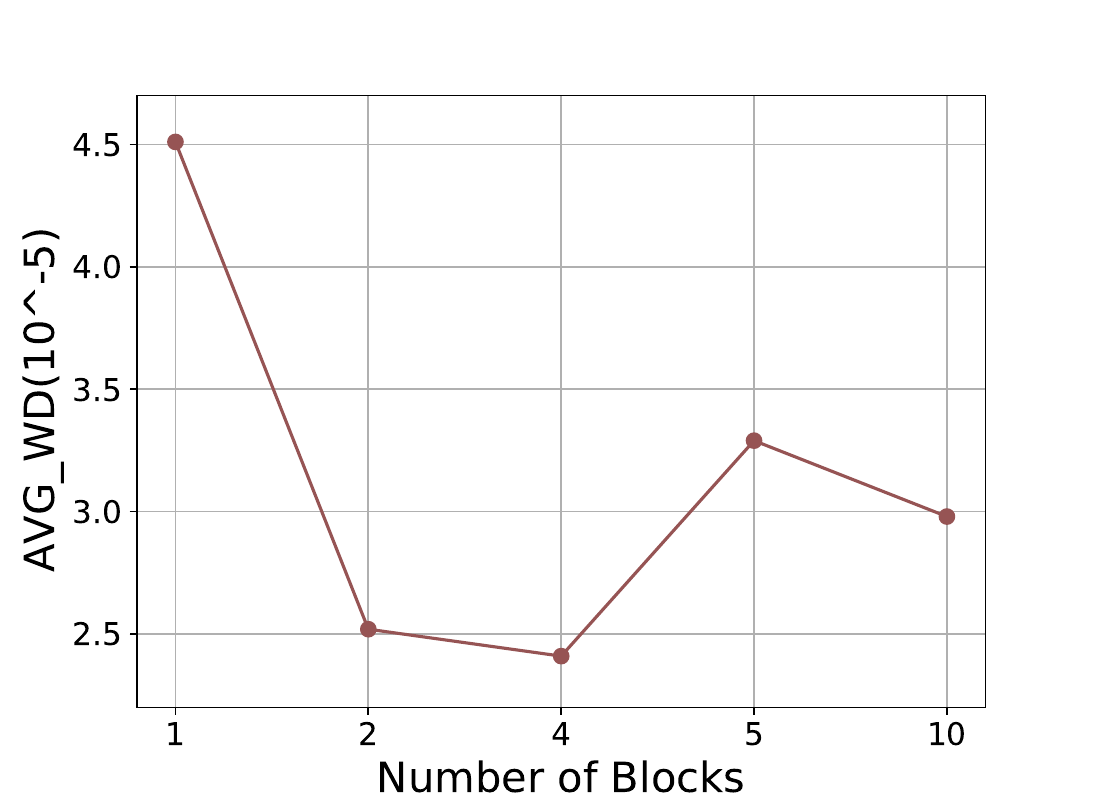}}
    \vspace{-3mm}
    \caption{Sensitivity test of number of attention heads.}
    \label{sen_2}
    \vspace{-7mm}
\end{figure}

To evaluate the effectiveness of each key component of the DSUF framework, we conduct ablation studies via diverse variants of the DSUF framework. We denote DSUF-A as the variant of the DSUF framework dropping the attention mechanism in configuration-level urban planning flows. DSUF-C is another variant without capturing geographical correlation via convolution extraction block. DSUF-ZP is the variant that drops the condition projection layers of zone-level urban planning flows. DSUF-LP is similar to DSUF-ZP by removing the autoregressive unconditional projection layers of configuration-level urban planning flows. We include more details. 
Figure \ref{abl} shows the results empirically that removing any component will decrease the performance of the DSUF framework which means every key component of the DSUF framework is effective.
\paragraph{Parameters Sensitivity Analysis.}
In this experiment, we are going to test whether our model is sensitive to hyper-parameters by analyzing the sensitivity of the number of blocks of configuration-level urban planning flows as well as the number of heads of multi-attentions. The result is indicated by Figure \ref{sen_2} and Figure \ref{sen}. We can observe that our framework is not sensitive to the number of blocks in configuration-level urban planning flows, since all metrics fluctuate in an acceptable interval.
In addition, except for only utilizing a single head, our framework is also not sensitive to the number of heads of multi-attentions.

\paragraph{Visualization of the generated land-use configurations.}

\begin{figure*}[t]
	\centering
    \subfigure[Original land-use configurations under different green level.]{
    \begin{minipage}[ht]{\linewidth}
    \centering
    \includegraphics[width=6.5in]{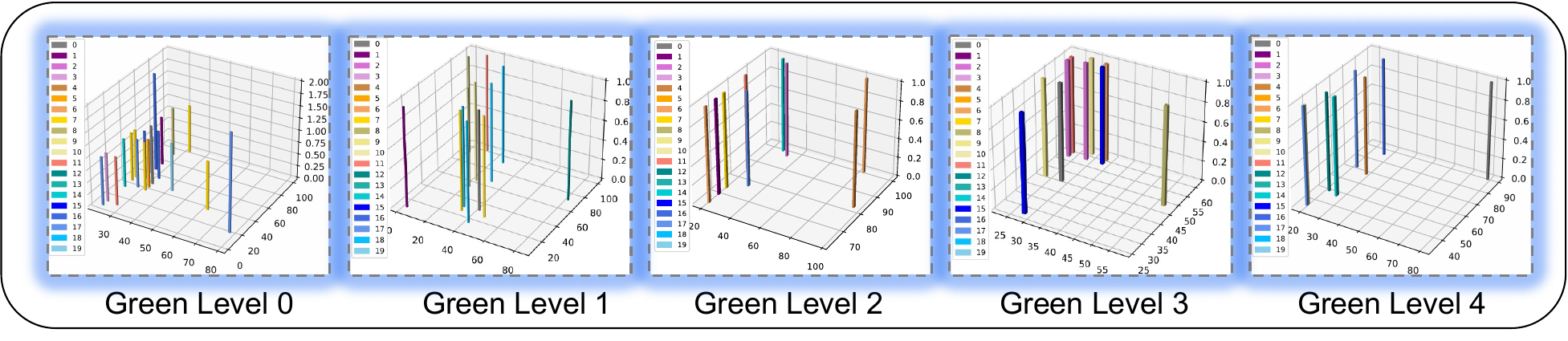}
    \end{minipage}%
    }%
    \\
    \vspace{-3mm}
    \subfigure[Generated land-use configurations by DSUF framework under different green level.]{
    \begin{minipage}[ht]{\linewidth}
    \centering
    \includegraphics[width=6.5in]{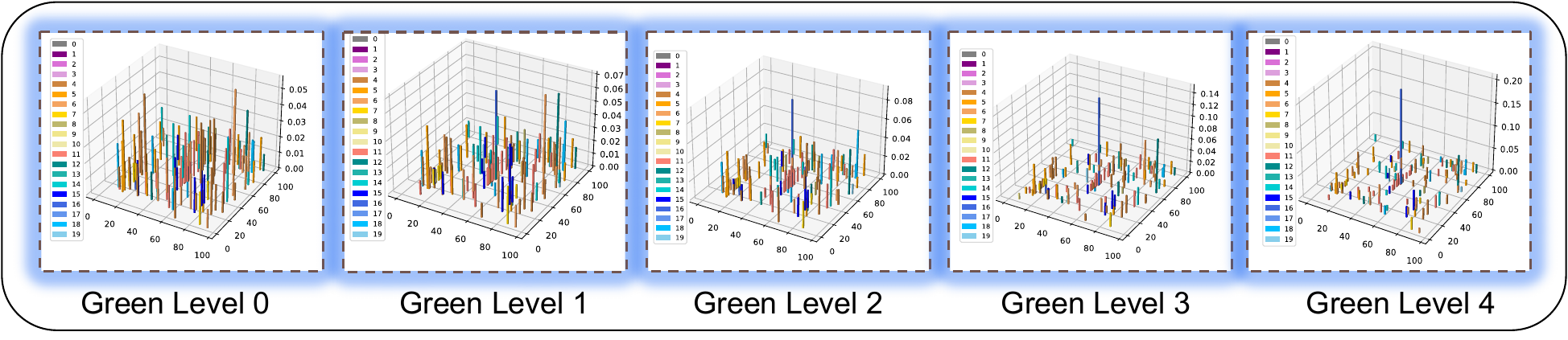}
    \end{minipage}%
    }%
    \vspace{-3mm}
    \centering
    \caption{Visualization of original and generated land-use configurations.}
    \label{generation}
\vspace{-5mm}
\end{figure*}

This case study aims to demonstrate how well our framework can understand human guidance and preserve the planning dependencies. Figure \ref{generation} indicates the visualizations of original and generated land-use configurations. From left to right, the green rate level increases monotonically and the distribution of POI becomes sparser which means there are more empty spaces available for greenery. 

Our generated results reflect the geographical information and semantic requirement which is feasible and meaningful for a modern city. Specifically, to satisfy the basic needs of white-collar workers, service-related POIs (e.g., POI category 4, 5, 6) are usually located near companies and financial institutions. 
Due to the high land price of the business urban functional zone, the number of industrial-related POIs (e.g., POI category 1, 2, 3) should be low.

\vspace{-0.3cm}
\section{Conclusion}
In this work, we propose the DSUF framework to generate land-use configurations automatically. Motivated by human reasoning to address intricate issues, we decompose the urban planning task into two stages. In the first stage, based on the surrounding contexts and human guidance, we generate urban functional zones as the rough sketch via zone-level urban planning flows. Consequently, we use the Information Fusion Module consisting of the semantic projection block and the convolution extraction block to capture the planning dependencies. In the second stage, we utilize Configuration-Level Urban Planning Flows to generate land-use configurations. 
Several empirical experiments indicate the supremacy and potency of DSUP Flows in the automatic urban planning task.

\bibliography{ref}
\bibliographystyle{unsrt}


\clearpage
\appendix
\section{Framework Details}
\subsection{Batch Normalization}
\label{batch_norm}
The batch normalization we use in our framework is to further ameliorate the propagation of signal. With batch normalization, we can not only improve the stability of training process, but also train a deeper stack of coupling layers to model more complex conditional distribution. The normalization procedure of zone-level urban planning flows can be written as: 
\begin{equation}
    \mathbf{U} \mapsto \frac{\mathbf{U}-\widetilde{\mu}_{\mathbf{U}}}{\sqrt{\widetilde{\sigma}_{\mathbf{U}}^2 +\epsilon}}
\end{equation}
Here, $\widetilde{\mu}_{\mathbf{U}}$ and $\widetilde{\sigma}_{\mathbf{U}}^2$ are the estimated batch statistics. The Jacobian determinant can be calculated as $\left(\prod_i \left( \widetilde{\sigma}_{\mathbf{U},i}^2 +\epsilon \right)\right)^{-\frac{1}{2}}$. The batch normalization used in configuration-level urban planning flows is similar.

\subsection{Inverse Transformation of each component}
\label{inverse_tran}
As mentioned in the main part of the paper, the design of flow-based networks requires all of the layers to be invertible. In this part, we give the inverse transformation of each flow layer we utilize. 
\paragraph{Conditional Coupling Layers} 
Conditional coupling layers are designed to apply effective learned function to estimate the conditional density of urban functional zones. Equation~\ref{ccl} indicates the forward transformation of conditional coupling layers. The inverse of them can be written as:
\begin{equation}
\begin{aligned}
\mathbf{h}_{\mathbf{1}}^{k} &= \mathbf{h}_{\mathbf{1}}^{k+1}\\
\mathbf{h}_{\mathbf{2}}^{k} &= exp(-f_{\theta, s}^k(\mathbf{h}_{\mathbf{1}}^{k+1} ;\mathbf{e}))\cdot (\mathbf{h}_{\mathbf{2}}^{k+1} - f_{\theta, b}^k(\mathbf{h}_{\mathbf{1}}^{k+1} ;\mathbf{e}))
\end{aligned}
\end{equation}

Here, the scale and bias are calculated by non-invertible fully connected layer $f_{\theta, s}^k$ and $f_{\theta, b}^k$.

\paragraph{Condition Projection Layers}
Condition projection layers aim to capture the knowledge of Urban Information Vector directly. With the partition, sparse input can be better processed via permutation without harming invertibility. The forward transformation of condition projection layers is revealed by Equation~\ref{cpl}, and the inverse transformation is:
\begin{equation}
\mathbf{h}_{\mathbf{2}}^{k} = exp(-f_{\theta, s}^k(\mathbf{e}))\cdot (\mathbf{h}_{\mathbf{2}}^{k+1} - f_{\theta, b}^k(\mathbf{e}))
\end{equation}
where the structure of $f_{\theta, s}^k$ and $f_{\theta, b}^k$ are similar to conditional coupling layers.

\paragraph{Conditional Autoregressive Layers}
We utilize conditional autoregressive layers to learn the distribution of land-use configuration conditional on the Attention matrix $\mathbf{A}$. The forward of conditional autoregressive layers in indicated by Equation~\ref{cal_forward}
We can formulate the inverse transformation as:
\begin{equation}
    \mathbf{m}_i^{k} =  exp(-g_{\theta^{\prime}, s_i}^k(\mathbf{m}_{1:i-1}^{k+1};\mathbf{A})) \cdot  (\mathbf{m}_i^{k+1}-g_{\theta^{\prime}, b_i}^k(\mathbf{m}_{1:i-1}^{k+1};\mathbf{A}))
\end{equation}
where $g_{\theta^{\prime}, s_i}^k$ and $g_{\theta^{\prime}, b_i}^k$ are functions which compute the mean and log standard deviation of the $i$-th element conditional on all previous variables and fused information embedding and the implementation of them is following the setting of MADE.

\paragraph{Autoregressive Unconditional Projection Layers}
To model the sparse information of land-use configurations, we employ autoregressive unconditional projection layers which has a forward transformation as Equation~\ref{aupl_forward}. And the inverse transformation can be written as:
\begin{equation}
    \mathbf{m}_i^{k} =  exp(-g_{\theta^{\prime}, s_i}^k(\mathbf{m}_{1:i-1}^{k+1})) \cdot  (\mathbf{m}_i^{k+1}-g_{\theta^{\prime}, b_i}^k(\mathbf{m}_{1:i-1}^{k+1}))
\end{equation}
where the construction of $g_{\theta^{\prime}, s_i}^k$ and $g_{\theta^{\prime}, b_i}^k$ is similar to conditional autoregressive layers.

\section{More Details of Experiments}
\subsection{More Details of Data Description}
\label{data_des}
\begin{table}[t]
 \scriptsize
\setlength{\tabcolsep}{1pt}
\caption{POI categories} 
\label{poi categories}
\begin{tabular}{c c c c c c}
\toprule[1pt]
code & POI category & code & POI category & code & POI category\\
\midrule
   0  & road & 1 & car service & 2 & car repair\\ 3 & motorbike service &
   4 & food service & 5 & shopping\\ 6 & daily life service & 7& recreation service&
   8& medical service \\ 9& lodging & 10& tourist attraction & 11 & real estate\\
   12 & government place & 13 & education & 14 & transportation\\15 &Finance&
   16 & company & 17 & road furniture\\18 & specific address & 19& public service\\
\bottomrule[1pt]

\end{tabular}
\end{table}
We show the 20 different POI categories which can can represent all fundamental configurations of modern cities in our dataset as Table~\ref{poi categories}.

\subsection{Calculation process of Evaluation Metrics}
\label{Eva_M}
To directly reflect the distribution distance between generated planning solutions and original configurations, we utilize three evaluation metrics to measure the performance under the setting of different green rate levels.
Specifically, 1) Average Kullback-Leibler (AVG$\_$KL) Divergence:
$\rm{AVG\_KL} = \frac{\sum_{j=1}^{5}w_j \cdot KL(X_j, \hat{X}_j)}{\sum_{j=1}^{5}w_j}$. 
2) Average Hellinger Distance (AVG$\_$HD):
$\rm{AVG\_HD}= \frac{\sum_{j=1}^{5}w_j \cdot HD(X_j, \hat{X}_j)}{\sum_{j=1}^{5}w_j}$.
3) Average Wasserstein Distance (AVG$\_$WD):
$\rm{AVG\_WD}= \frac{\sum_{j=1}^{5}w_j \cdot WD(X_j, \hat{X}_j)}{\sum_{j=1}^{5}w_j}$.
Here, we denote $w_j$ as the number of land-use configurations of human guidance level $j$, $X_j$ as the distribution of original land-use configurations of human guidance level $j$, $\hat{X}_j$ as the distribution of generated land-use configurations of human guidance level $j$.

\end{document}